\documentclass[conference]{IEEEtran}
\IEEEoverridecommandlockouts

\usepackage{cite}
\usepackage{url}
\usepackage{amsmath,amssymb,amsfonts}
\usepackage{algorithm}
\usepackage{algorithmic}
\usepackage{graphicx}
\usepackage{tikz}
\usepackage{textcomp}
\usepackage{xcolor}
\usepackage{booktabs}
\usepackage{multirow}
\def\BibTeX{{\rm B\kern-.05em{\sc i\kern-.025em b}\kern-.08em
    T\kern-.1667em\lower.7ex\hbox{E}\kern-.125emX}}
\begin{document}

\title{Comparing statistical and deep learning techniques for parameter estimation of continuous-time stochastic differentiable equations.\\
}

\author{\IEEEauthorblockN{1\textsuperscript{st} Aroon Sankoh}
\IEEEauthorblockA{\textit{Department of Computer Science and Engineering} \\
\textit{Washington University in St. Louis}\\
St. Louis, United States \\
a.j.sankoh@wustl.edu}
\and
\IEEEauthorblockN{2\textsuperscript{nd} Mladen Victor Wickerhauser}
\IEEEauthorblockA{\textit{Department of Mathematics} \\
\textit{Washington University in St. Louis}\\
St. Louis, United States \\
victor@wustl.edu}
}

\maketitle
\thispagestyle{plain}
\pagestyle{plain}

\setlength{\parskip}{0pt}

\begin{abstract}
Stochastic differential equations such as the Ornstein-Uhlenbeck process have long been used to model real-world probablistic events such as stock prices and temperature fluctuations. While statistical methods such as Maximum Likelihood Estimation (MLE), Kalman Filtering, Inverse Variable Method, and more have historically been used to estimate the parameters of stochastic differential equations, the recent explosion of deep learning technology suggests that models such as a Recurrent Neural Network (RNN) could produce more precise estimators. We present a series of experiments that compare the estimation accuracy and computational expensiveness of a statistical method (MLE) with a deep learning model (RNN) for the parameters of the Ornstein-Uhlenbeck process. 
\end{abstract}

\section{Introduction}
In section I, we will define the Ornstein-Uhlenbeck (OU) stochastic process and explore some of the theory behind its solution. After introducing useful properties of the OU process, we can define the likelihood function to optimize for MLE estimation and search algorithm(s) we will use to obtain estimates. Section II discusses the network architecture for our recurrent neural network and experimental settings used in training and testing. Section III describes the results of both parameter estimation techniques and interprets them in the context of relevant works. Section IV summarizes and concludes our case study.

\subsection{Processes of Ornstein-Uhlenbeck Type}

A process of $X = \{X_t\}$ is Ornstein-Uhlenbeck (OU) if it satisfies the following stochastic differential equation: 

\begin{equation}
    \begin{aligned}
        dX_t = -\theta X_tdt + \sigma dW_t \\
        X_0 = X_0, t \geq 0 
    \end{aligned}
\end{equation}

with $(\theta, \sigma) \in (0, \infty) \times (0, \infty)$ and $t_i = i\Delta$ for $i = 1,2,...$ and fixed increments $\Delta > 0$. $W_t$ is standard Brownian Motion. 
We can find a unique, strong solution to $(1)$ using Itô's Formula. Consider the integrating factor: 
\begin{equation*}
I_t = e^{\theta t}
\end{equation*}
Multiplying both sides of $(1)$ by $I_t$ leads to: 
\begin{equation*}
e^{\theta t}dX_t = -e^{\theta t}X_tdt + \sigma e^{\theta t}d W_t
\end{equation*}
Rewriting the left-hand-side using the product rule: 
\begin{equation*}
d(e^{\theta t}X_t) = \sigma e^{{\theta t}}dW_t
\end{equation*}
Integrating from $0$ to $t$:
\begin{equation*}
e^{\theta t}X_t = X_0 + \sigma\int_{o}^{t} e^{{\theta s}}dW_s
\end{equation*}
Leaves us with the solution: 
\begin{equation}
X_t = X_0e^{-\theta t} + \sigma\int_{o}^{t} e^{{-\theta (t-s)}}dW_s
\end{equation}
The existence of $(2)$ also means that $(1)$ is well-defined. See [1] for more details. 

It is well-known that if $X_0 \sim N(0, \frac{\sigma^2}{2\theta})$ then (2) is a time-stationary Gauss-Markov process with continuous sample paths [2]; we can easily confirm these properties below: 
\begin{enumerate}
    \item \textbf{Gaussian.} $X_0e^{-\theta t}$ is deterministic $(X_0 = X_0)$ and $\sigma\int_{o}^{t} e^{{-\theta (t-s)}}dW_s$ is Gaussian as it is a Weiner integral with respect to Brownian Motion $W_t$, which is a Gaussian process itself. Any linear combination of a deterministic and Gaussian process is Gaussian.
    \item \textbf{Time-Stationary.} (2) has a bounded constant mean and its' auto-covariance function only depends on time increments $|t-s|$, not on $t$ or $s$ seperatedly [1].
    \item \textbf{Markovian.} A markov process $M_t$ is one where the future state of the process only depends on its' present state and not any past states: $P(M_t| M_s \cap M_r) = P(M_t| M_s)$ where $r < s$. By examining the explicit solution: 
    \begin{equation*}
        X_t = X_se^{-\theta (t-s)} + \sigma\int_{s}^{t} e^{{-\theta (t-r)}}dW_r
    \end{equation*}
    We see that the first term only depends on $s$ (the previous state) and the second term is independent of $X_{<s}$ given $X_s$ since it depends on future increments of $W_t$, and Brownian motion has independent increments.
\end{enumerate}

Note that $X_t$ has the following mean, variance, and covariance functions: 
\begin{equation*}
        E(X_t) = E(X_0)e^{-\theta t}
\end{equation*}
\begin{equation*}
        Var(X_t) = \frac{\sigma^2}{2\theta} + e^{-2\theta t} \left(Var(X_0) - \frac{\sigma^2}{2\theta}\right)
\end{equation*}
\begin{equation*}
        C(t, s) = \frac{\sigma^2}{2\theta}e^{-\theta |t - s|}
\end{equation*}

We can use the Gaussian and Markovian properties, along with the known transition density of the Ornstein Uhlenbeck process to derive the likelihood function for the parameter vector $\eta = (\sigma^2, \theta)$. Doob [2] first dervied the transition density, or the probability of a process being in state $x$ at time $t$ given the process was in state $x'$ at time $s$, using the Markov property. 
\begin{align*}
    P(X_t = x \mid X_s = x') 
        &= \frac{1}{\sqrt{2\pi V(t-s)}} \\
        &\quad \times \exp\left(-\frac{\left(x-x'e^{-\theta (t-s)}\right)^2}{2V(t-s)}\right)
\end{align*}

where $V(t-s) = \frac{\sigma^2}{2\theta}\left( 1-e^{-2\theta(t-s)}\right)$, this term accounts for the dispersion of the process over time, dependent on the relationship between $\theta$ and $\sigma^2$. 

Suppose $X = \{X_{t_o}, X_{t_1}, X_{t_2}, . .. X_{t_n}\} $ is a sequence of $n+1$ observations of our process described by $(2)$. The likelihood function for the parameter $\eta$ is constructed as the product of the transition densities between consecutive observations: 
\begin{align*}
    L(\eta) &= \prod_{i=1}^{n} P(X_{t_i} |X_{t_{i-1}}) \\ 
    \hspace{-3cm} & = \prod_{i=1}^{n} \frac{1}{\sqrt{2\pi V(t_i - t_{i-1})}} \\ 
    \hspace{-3cm} & \times \exp \left(\frac{\left(X_{t_i}-X_{t_{i-1}}e^{-\theta (t_i-t_{i-1})}\right)^2}{2V(t_i-t_{i-1})}\right)
\end{align*}

Now if we let $\Delta t = t_i-t_{i-1}$, the log-likelihood becomes: 
\begin{align*}
    \log (L(\eta)) &= -\frac{1}{2} \sum_{i=1}^{n} \left[ \log(2\pi V(\Delta t_i)) \right. \\
    &\quad \left. + \frac{\left( X_{t_i} - X_{t_{i-1}} e^{-\theta \Delta t_i} \right)^2}{V(\Delta t_i)} \right]  
\end{align*}

Plugging $V(\Delta t_i) = \frac{\sigma^2}{2\theta}(1-e^{-2 \theta(\Delta t_i)})$ solves to our final log-likelihood function: 
\begin{equation}
\begin{aligned}
    \log (L(\eta)) &= \frac{n}{2}\log(2\pi) \\ 
    & -\frac{1}{2} \sum_{i=1}^{n}\log\left(\frac{\sigma^2}{2\theta}(1-e^{-2 \theta\Delta t_i})\right)  \\
    & - \frac{1}{2} \sum_{i=1}^{n}\frac{\left(X_{t_i}-X_{t_{i-1}}e^{-\theta \Delta t_i}\right)^2}{\frac{\sigma^2}{2\theta}(1-e^{-2 \theta(\Delta t_i)})}
\end{aligned}
\end{equation}

Iacus [3] showed that if we take the maximum of (3) we can find a closed form solution for $\sigma^2$. But since $\theta$ must be estimated numerically, we will estimate $\eta$ numerically.

\subsection{Optimization Procedure}
Many optimization techniques for parameter estimation of stochastic differential equations have been proposed. Valdivieso, Schoutens, and Tuerlinckx [4] used a modified global differential evolution algorithm developed by [5] to estimate parameters of one-dimensional Ornstein-Uhlenbeck processes driven by a background Levy process. Markov Chain Monte Carlo methods have been proposed by [6] to estimate parameters for diffusion processes, specifically for latent variable models. Durham and Gallant [7] propose simulation based MLE techniques that use Monte Carlo techniques to improve on [8]'s seminal simulation scheme. Various method of moments (MOM) estimators have also been proposed, Gallant and Tauchen [9] introduced an efficient MOM estimator to generate ideal generalized MOM conditions for MLE. 

We propose a hybrid approach that \textbf{I.} Uses generalized method of moments for parameter initialization. \textbf{II.} Refines the parameter estimates using the Broyden-Fletcher-Goldfarb-Shanno (BFGS) local optimization algorithm. \textbf{III.} Employs a basin-hopping global optimization algorithm if the likelihood surface remains extremely non-convex after parameter refinement. Pseudo-code for the generalized method of moments and BFGS algorithms can be found below. In this paper, we used Python version \texttt{3.12.9} for all algorithm implementations, simulations, and statistical analysis. 

\begin{algorithm}
\caption{Generalized Method of Moments}
\begin{algorithmic}[1]
    \STATE \textbf{procedure} \textsc{GMM}$(X, \Delta t)$
    \STATE $X \leftarrow \text{array}(X)$
    \STATE $\hat{\rho} \leftarrow \text{corrcoef}(X_{1:T-1}, X_{2:T})_{1,2}$
    \STATE $\hat{\rho} \leftarrow \min(\max(\hat{\rho}, 10^{-4}),\ 0.999)$
    \STATE $\hat{\theta} \leftarrow \max\left( \frac{-\log(\hat{\rho})}{\Delta t},\ 0.5 \right)$
    \STATE $\hat{\sigma}^2 \leftarrow 2 \cdot \hat{\theta} \cdot \text{Var}(X)$
    \STATE \textbf{return} $(\hat{\theta}, \hat{\sigma}^2)$
\end{algorithmic}
\end{algorithm}

\begin{algorithm}
\caption{Broyden–Fletcher–Goldfarb–Shanno}
\begin{algorithmic}[1]
    \STATE \textbf{procedure} \textsc{BFGS}$(f, \nabla f, x_0, H_0, \epsilon)$
    \STATE $x \leftarrow x_0$
    \STATE $H \leftarrow H_0$
    \REPEAT
        \STATE $g \leftarrow \nabla f(x)$
        \IF{$\|g\| < \epsilon$}
            \STATE \textbf{return} $x$
        \ENDIF
        \STATE $p \leftarrow -H g$ \hfill 
        \STATE $\alpha \leftarrow \text{LineSearch}(f, x, p)$
        \STATE $x_{\text{new}} \leftarrow x + \alpha p$
        \STATE $s \leftarrow x_{\text{new}} - x$
        \STATE $y \leftarrow \nabla f(x_{\text{new}}) - g$
        \STATE $\rho \leftarrow \frac{1}{y^\top s}$
        \STATE $H \leftarrow (I - \rho s y^\top) H (I - \rho y s^\top) + \rho s s^\top$
        \STATE $x \leftarrow x_{\text{new}}$
    \UNTIL{convergence}
    \STATE \textbf{return} $x$
\end{algorithmic}
\end{algorithm}

\section{Deep Learning Parameter Estimators}
The exponential increase in computing power has catalzyed the wide-spread application of deep learning methods to the natural sciences, medical fields, and finance over the past twenty years. Stochastic processes are used to model random events within all these areas, suggesting that deep learning models can accurately estimate the parameters of such processes. Previous efforts at parameter estimation via deep learning focus on partial differential equations [10-11], or investigating trajectories in anomalous diffusion [12-13]. 

Works that explore parameter estimation of specific stochastic processes via deep learning [14] often do so with trajectory-rich training sets and high-performance GPUs. Since typically only research labs and companies have access to the most powerful graphics cards, a lightweight neural network that can be trained and tested using minimal resources will provide the best alternative to traditional estimation techniques for independent researchers. We propose a Long-Short-Term-Memory (LSTM) neural network that can be trained on as few as 20,000 sample paths for the estimation of the drift and volatility parameters of the Ornstein-Uhlenbeck process. 

\subsection{Network Architecture}

Long-Short-Term-Memory models are a type of recurrent neural network (RNN) specifically proposed [15] to address the vanishing gradient problem that is commonly encountered by traditional RNNs. They excel at capturing long-term dependencies in sequential data, thus making them well-suited for tasks such as speech recognition, natural language processing, and parameter estimation of discrete-time processes. Feng et al. [16] used an LSTM coupled with a fully connected neural network to estimate the parameters of stochastic differential equations driven by fractional Brownian motion. Our LSTM 2-layer network structure is depicted in Figure 1; its purpose is to extract deep features $h_t$ from a trajectory $X_t$. It does so by analyzing the temporal structure of an input sample path of 500 time steps: $\{X_0, X_1, X_2, ...X_{499}\}$ and encoding it into a fixed length vector $\vec{h_t}$ (the hidden state).

\begin{figure}[h!]
    \centering
    \includegraphics[width=0.5\textwidth]{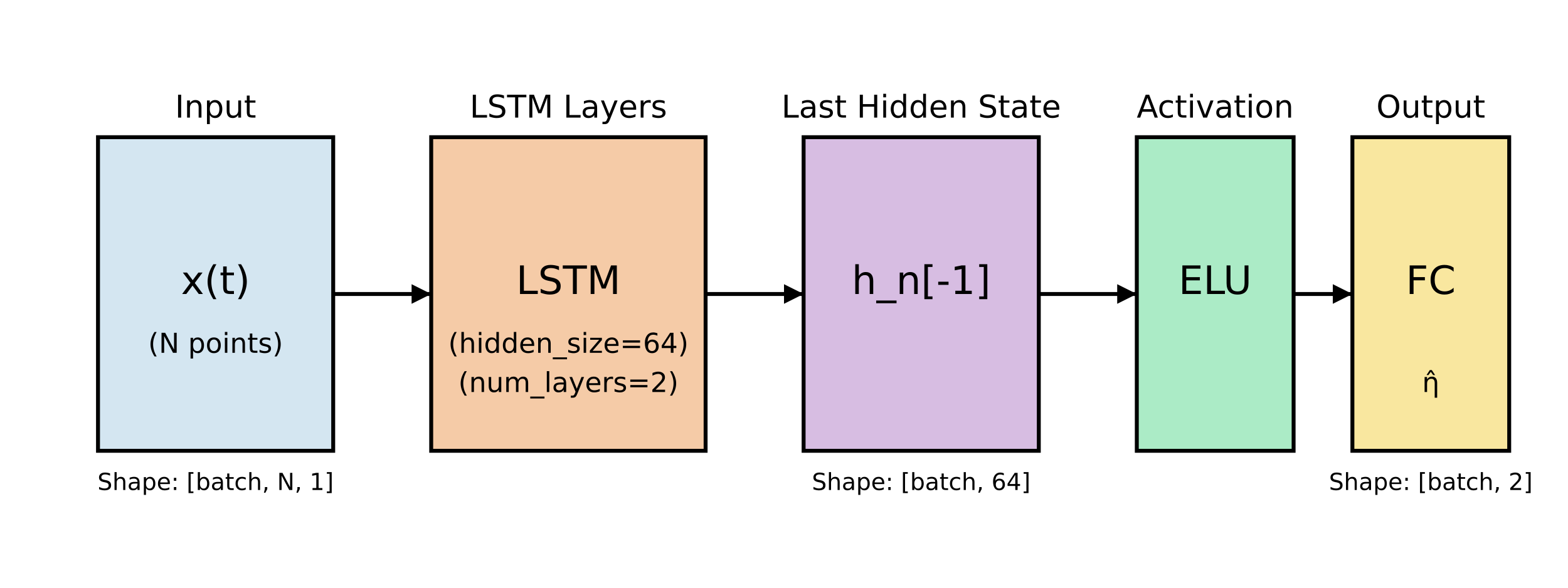}
    \caption{The full model architecture.}
    \label{fig:loss_curve}
\end{figure}

The exponential linear unit activation function (ELU) is then applied to $h_t$ before it is fed into fully connectd output layer that maps to our parameter vector $\hat\eta$. The complete function mapping of the LSTM layer is below: 
\begin{align*}
    i_t &= \sigma(W_{ii} x_t + b_{ii} + W_{hi} h_{t-1} + b_{hi}) \\ 
    f_t &= \sigma(W_{if} x_t + b_{if} + W_{hf} h_{t-1} + b_{hf}) \\
    g_t &= \tanh(W_{ig} x_t + b_{ig} + W_{hg} h_{t-1} + b_{hg}) \\
    o_t &= \sigma(W_{io} x_t + b_{io} + W_{ho} h_{t-1} + b_{ho}) \\
    c_t &= f_t \odot c_{t-1} + i_t \odot g_t \\
    h_t &= o_t \odot \tanh(c_t)
\end{align*}
where $i_t, f_t, g_t, o_t$ are the input, forget, cell, and output gates, respectively, $c_t$ is the cell state at time $t$, and $h_t$ is the hidden state at time $t$. For each gate $\in\{i, f, g, o\}$ at time $t$, $W_{ii}$ is the input-to-input gate weight matrix, $W_{hi}$ is the hidden-to-input gate weight matrix, etc., while $b_{ii}$ is the input bias for the input gate, $b_{hi}$ is the hidden bias for the input gate, and so on. $\odot$ is the element-wise product and $\sigma$ is the sigmoid function.

Our chosen \textit{loss function} will be the Huber Loss function. Huber is a piecewise loss function that amalgamates the strengths of Mean Squared Error (MSR), a loss that is sensitive to small errors, and Mean Absolute Error (MAE), a loss that reduces sensitivity to outliers. Kumar [17] used the Huber loss in a deep learning based parameter estimation experiment to account for noise and corruptions in training data. 

Given a prediction $\hat{y}$, a target value $y$, and the residual error $\hat{y}-y$, the Huber loss $L_\delta (r)$ is defined as: 
\begin{equation}
L_\delta(r) = 
\begin{cases}
    \frac{1}{2} r^2 & \text{if } |r| \leq \delta \\
    \delta (|r| - \frac{1}{2} \delta) & \text{if } |r| > \delta
\end{cases}
\end{equation}

$\delta$ controls when $L_\delta (r)$ behaves like MSE versus MAE. If error is small $(|r| \leq \delta)$, then the loss is quadratic and behaves more like MSE. If the error is large $(|r| > \delta)$, then the loss is linear and behaves more like MAE. Since our training data is noise-free but we still want to preserve robustness against outliers, we chose $\delta = 1$ to balance these two characteristics.
We apply the Huber loss independently to each of the two output parameters $(\hat\theta, \hat\sigma^2)$. However, since these two parameters often differ in scale and in their sensitivity to estimation errors, we introduce a weighted composite loss function (5) to account for this imbalance. 

\begin{equation}
\begin{aligned}
L_{\text{total}} &= w_\theta \cdot L_\delta(\hat{\theta}, \theta)  + w_{\sigma^2} \cdot L_\delta(\hat{\sigma}^2, \sigma^2)
\end{aligned}
\end{equation}

We selected $w_\theta = 1$ and $w_{\sigma^2} = 0.5$ to account for the greater consistency in predicting $\hat\theta$ observed during training and the wider variance in $\hat{\sigma}$.

Our chosen \textit{activation function} is the Exponential Linear Unit (ELU) function [18]. ELUs are similar to ReLUs and its variants in that they alleviate the vanishing gradient problem. But in contrast to ReLUs, which are strictly nonnegative, ELUs include negative values that allow them to push mean unit activations closer to zero. This corrects for the bias shift for units between layers, and leads to faster learning, improved learning stability, and potentially improved classification accuracy. The exponential linear unit (ELU) with $0 < \alpha$ is: 

\begin{equation}
f_\alpha(x) =
\begin{cases}
x & \text{if } x > 0 \\
\alpha (e^x - 1) & \text{if } x \leq 0
\end{cases}
\end{equation}

where the hyperparameter $\alpha$ controls the saturation value for negative inputs [18]. So when the unit input is positive $(x>0)$, $f_\alpha(x)$ behaves like ReLU and returns $x$. Otherwise, $f_\alpha(x)$ smoothly decays to $-\infty$ as $x \rightarrow \infty$. 

\subsection{Experimental Settings and Hyperparameters}
The network has 2 stacked LSTM layers, 1 ELU layer, and 1 fully-connected output layer. The model is implemented in PyTorch and trained and tested on only a 2 GB GPU with 6 GB system RAM and 2 CPU cores to ensure accessibility and reproducibility by independent researchers who lack access to high performance GPUs. The parameter vector to be estimated $\eta$ remains the same.  

Our model is trained for $100$ epochs with a batch size of $128$ and learning rate of $.001$. The input trajectories are normalized for faster convergence. To maintain consistency between methods, the model is trained on $5000$ trajectories of each of the $4$ parameter combinations in Table 1. We split the input dataset of $20,000$ total trajectories $80/20$ into training and validation sets to allow validation loss tracking per epoch. The network is optimized via ADAM, and trained using mini-batch stochastic gradient descent.  

\section{Results}

\subsection{MLE $\hat\eta$ Estimates}
True parameter values for $\eta$ can be found in Table 1 below. We will use the stationary variance of an OU process, $\frac{\sigma^2}{2 \theta}$, as an uncertainty parameter for our estimator $\hat{\theta}$. That is, if $\sigma >> \theta$, we are highly uncertain of the accuracy of our estimator $\hat{\theta}$ because the high volatility impedes observation of mean reversion. But if, $\sigma << \theta$, we are not uncertain of $\hat{\theta}$ since the mean reversion can easily be observed in simulated sample paths.

We also found that varying the initial starting point of each path $X_0$ by a multiple of $\sigma$ greatly improves the accuracy of our $\hat\theta$. This is due to the drift caused by $\theta$ being proportional to the distance from the mean; 0 in this case. As $X_t$ increases, the mean reversion term in the OU process $(-\theta X_tdt)$ dominates the diffusion component $(\sigma dW_t)$. Our simulation uniformly samples initial values from a range of $[-k\sigma, k\sigma]$. We selected the multiple $k=30$ but repeated simulations show that $k \geq 10$ will suffice. 

\begin{table}[htbp]
    \centering
    \caption{True Parameter Values}
    \begin{tabular}{|c|c|c|c|}
        \hline
        \textbf{Parameter Combinations} & $\theta$ & $\sigma^2$ & $\sigma^2/ 2\theta$ \\ \hline
        Strong Mean Reversion & 2.0 & 1.0 & 0.25\\ \hline
        Weak Mean Reversion & 0.2 & 1.0 & 2.5 \\ \hline
        High Volatility & 0.5 & 4.0 & 4.0 \\ \hline
        Low Volatility & 0.5 & 0.25 & 0.0625 \\ \hline
    \end{tabular}
    \label{tab:empty}
\end{table}

First attempts at calculating $\hat\eta$ using just BFGS yielded unreliable results, so we employed the global basin-hopping algorithm to evaluate BFGS at multiple randomly selected coordinates. Our optimization method took $115.5$ seconds on average to converge on $\hat\eta$. Table 2 depicts the statistics of $\hat\eta$ for MLE. 

The log-likelihood function evaluation and basin-hopping algorithm took the majority of time and CPU usage to compute. On average, it took $108$ seconds and $531.375$ MB of GPU memory to complete the optimization of the simulated paths. Time and memory profiling were completed with the Python package Scalene version \texttt{1.5.51}. 

\subsection{RNN $\hat\eta$ Estimates}

\begin{figure}[h!]
    \centering
    \includegraphics[width=0.5\textwidth]{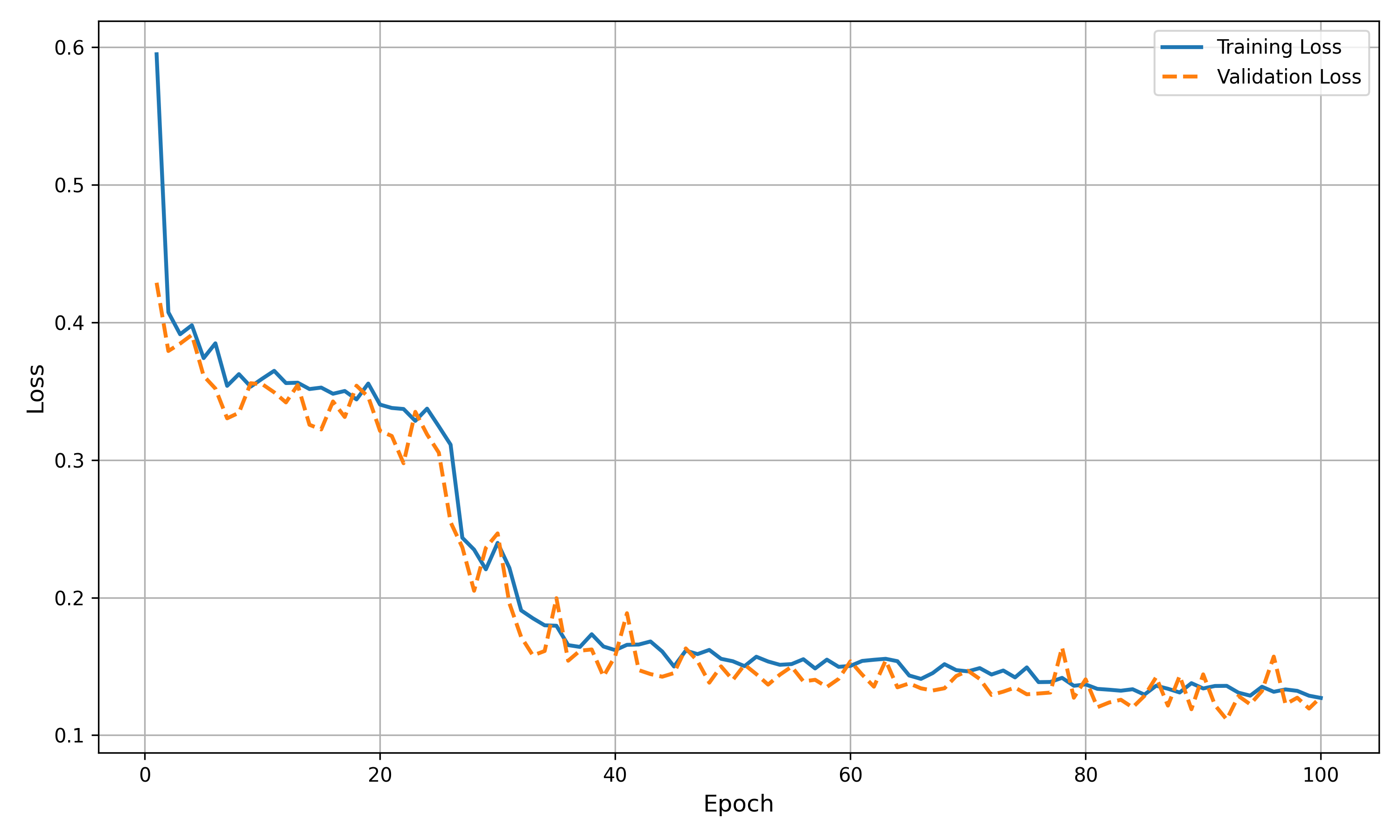}
    \caption{Training and validation loss over 100 epochs.}
    \label{fig:loss_curve}
\end{figure}

\begin{table*}[ht]
\centering
\caption{MLE and RNN Estimation Statistics Across Parameter Regimes}
\renewcommand{\arraystretch}{1.5}
\setlength{\tabcolsep}{9pt}
\begin{tabular}{|c|c|c||cccc||cccc|}
\hline
\textbf{True $\theta$} & \textbf{True $\sigma^2$} & \textbf{Estimator} 
& \multicolumn{4}{c||}{$\hat\theta$} 
& \multicolumn{4}{c|}{$\hat{\sigma}^2$} \\
\cline{4-11}
& & & Mean & Median & Std & RMSE & Mean & Median & Std & RMSE \\
\hline

\multirow{2}{*}{0.2 (Weak)} & \multirow{2}{*}{1.0} 
& MLE & 0.2883 & \textbf{0.2033} & 0.5160 & 0.5234 & \textbf{0.9981} & 0.9971 & \textbf{0.0643} & \textbf{0.0643} \\
& & RNN & \textbf{0.2547} & 0.2130 & \textbf{0.0861} & \textbf{0.1020} & 0.9517 & \textbf{1.0005} & 0.1977 & 0.2035 \\
\hline

\multirow{2}{*}{2.0 (Strong)} & \multirow{2}{*}{1.0} 
& MLE & \textbf{2.0927} & \textbf{2.0086} & 0.6201 & 0.6269 & \textbf{1.0017} & \textbf{1.0004} & \textbf{0.0646} & \textbf{0.0646} \\
& & RNN & 1.8819 & 1.9665 & \textbf{0.3024} & \textbf{0.3247} & 0.9928 & 1.0120 & 0.0785 & 0.0788 \\
\hline

\multirow{2}{*}{0.5} & \multirow{2}{*}{0.25 (Low)} 
& MLE & 0.5868 & \textbf{0.5044} & 0.5388 & 0.5457 & \textbf{0.2497} & \textbf{0.2494} & \textbf{0.0159} & \textbf{0.0159} \\
& & RNN & \textbf{0.4803} & 0.4496 & \textbf{0.1445} & \textbf{0.1458} & 0.5836 & 0.4730 & 0.4272 & 0.5420 \\
\hline

\multirow{2}{*}{0.5} & \multirow{2}{*}{4.00 (High)} 
& MLE & 0.5868 & \textbf{0.5044} & 0.5388 & 0.5457 & \textbf{3.9948} & \textbf{3.9904 }& \textbf{0.2582} & \textbf{0.2582} \\
& & RNN & \textbf{0.5104} & 0.4834 & \textbf{0.3117} & \textbf{0.3118} & 3.0268 & 3.9708 & 1.4207 & 1.7221 \\
\hline
\end{tabular}
\vspace{0.5em}\\ 
\label{tab:rnn_mle_comparison}
\end{table*}

Training the model took only 2 hours and 15 minutes using the GPU described in section II. The training and validation loss curves depicted in Figure 2 show model convergence. 281 MB of memory were used on average, with both the forward and backward propagation functions consuming 1 GB of GPU memory during their execution. Unsurprisingly, forward propagation accounted for $60$ percent of the programs run time. Model inference is performed on 500 sample trajectories generated with strictly the parameter combinations in Table 1. Although one of the benefits of deep learning models is their ability to classify inputs not seen during training, we neglect to test the generalizability of our RNN since our focus is to make comparisons with statistical estimation techniques. Inference took 8 seconds and used 63 MB of memory on average. Table 2 depicts the statistics of $\hat\eta$ for RNN.  

\subsection{Interpretation}
Statistics for the true $\eta$ values of $(2,1)$, $(0.2,1)$ are negligible between the two estimation methods. Marginally, the RNN produced $\hat\theta$ estimates with tighter precision but less accuracy, while the MLE produced $\hat\sigma^2$ estimates with tighter precision and comparable accuracy. Statistics for the true $\eta$ values of $(0.5, 4)$, $ (0.5, 0.25)$ are where the results between the two techniques see noticeable variation. 

While the $\hat\theta$ estimates produced by the RNN were marginally more precise and accurate, $\hat\sigma^2$ of the RNN showed significant error compared to MLE. MLE is more stable in both high and low volatility regimes whereas RNN not only exhibits high standard deviation—leading to poor precision—but also biased mean estimates. RNN excels at learning nonlinear mappings for $\theta$, even when the underlying distribution (Gaussian in our case), is noisy or has high variability. Previous works have shown that there is a limit to the amount of sample trajectories and layers you can add to a deep learning model before parameter estimation accuracy plateaus [14]. But RNN struggles to predict $\sigma^2$ for trajectories with high or low volatility. MLE remains the gold standard for volatility estimation, especially in well-specified models. 

Considering the small size and homogeneity of the training set, our deep learning model performed exceptionally well in calculating $\hat\theta$. Higher quality datasets train higher quality models, so we recommend future researchers train an RNN on more sample trajectories with thousands of potential parameter combinations to see further improvements in the estimation of both $\theta$ and $\sigma^2$. Traditional parameter estimation techniques such as MLE exhibit higher invariance to the input data; increasing the amount of sample trajectories used in optimization had no noticeable affect on the accuracy of the estimates. 

Since MLE is only optimal under the correct specifications (i.e. functional form, distribution, and independence structure), this suggests that deep learning models are superior estimators for abundant amounts of real-world data. Although our trajectories were simulated using the closed-form solution in (2), deep learning models are particularly advantageous in settings with abundant, high-dimensional, or noisy data drawn from unknown or complex distributions—such as spatiotemporal fluctuations or financial markets. In such settings, the flexibility of neural networks to learn directly from data without requiring apriori assumptions is critical. Future work can explore this conjecture further.

\section{Conclusion}

We have completed a thorough investigation into the viability of various parameter estimation techniques for the Ornstein-Uhlenbeck stochastic process. Our results show that a light-weight RNN trained on minimal sample trajectories perform marginally better for the estimation of the mean-reverting term $\theta$, while MLE is significantly better at estimating the volatility term $\sigma^2$ for high/low volatility processes. These results suggest that deep learning models can serve as superior alternatives to traditional techniques in scenarios where model assumptions are uncertain or data is abundant. Future efforts should explore a broader range of model architectures tailored for parameter estimation, especially in real world settings where the underlying distribution is unknown.

\section*{Acknowledgment}

The authors appreciate the open source contribution of the Aleatory package (https://github.com/quantgirluk/aleatory), which provided the core functionality for sample generation in this work. The authors would also like to thank the Department of Computer Science and Engineering at Washington University in St. Louis for providing GPU resources that aided computational work done for this research. This research did not receive any specific grant from funding agencies in the public, commercial, or not-for-profit sectors.

\end{document}